\documentclass[10pt,twocolumn,letterpaper]{article}

\usepackage{iccv}
\usepackage{times}
\usepackage{epsfig}
\usepackage{graphicx}
\usepackage{amsmath}
\usepackage{amssymb}
\usepackage{booktabs}
\usepackage{diagbox} 
\usepackage{multirow}
\iccvfinalcopy
\usepackage[breaklinks=true,bookmarks=false]{hyperref}
\def\iccvPaperID{11661} % *** Enter the ICCV Paper ID here

\ificcvfinal\fi

\begin{document}

\title{Multi-scale Geometry-aware Transformer for 3D Point Cloud Classification}

\author{Xian Wei\\
TUM\\
{\tt\small xian.wei@tum.de}
\and
Muyu Wang\\
Liaoning Technical University\\
{\tt\small 598477187@qq.com}
\and
Shing-Ho Jonathan Lin\\
University of Chinese Academy of Sciences\\
{\tt\small linchenghao21@mails.ucas.ac.cn}
\and
Zhengyu Li\\
East China Normal University\\
{\tt\small lzy1999729@gmail.com}
\and
Jian Yang\\
Information Engineering University\\
{\tt\small jian.yang@tum.de}
\and
Arafat Al-Jawari\\
University of Chinese Academy of Sciences\\
{\tt\small arafatmmj@gmail.com}
\and
XUAN TANG\\
East China Normal University\\
{\tt\small 2265275624@qq.com}
}
\maketitle

%%%%%%%%% ABSTRACT
\begin{abstract}
 
 Self-attention modules have demonstrated remarkable capabilities in capturing long-range relationships and improving the performance of point cloud tasks. However, point cloud objects are typically characterized by complex, disordered, and non-Euclidean spatial structures with multiple scales, and their behavior is often dynamic and unpredictable. The current self-attention modules mostly rely on dot product multiplication and dimension alignment among query-key-value features, which cannot adequately capture the multi-scale non-Euclidean structures of point cloud objects. To address these problems, this paper proposes a self-attention plug-in module with its variants, Multi-scale Geometry-aware Transformer (\emph{MGT}). \emph{MGT} processes point cloud data with multi-scale local and global geometric information in the following three aspects. At first, the \emph{MGT} divides point cloud data into patches with multiple scales.  Secondly, a local feature extractor based on sphere mapping is proposed to explore the geometry inner each patch and generate a fixed-length representation for each patch. Thirdly, the fixed-length representations are fed into a novel geodesic-based self-attention to capture the global non-Euclidean geometry between patches. Finally, all the modules are integrated into the framework of \emph{MGT} with an end-to-end training scheme. Experimental results demonstrate that the \emph{MGT} vastly increases the capability of capturing multi-scale geometry using the self-attention mechanism and achieves strong competitive performance on mainstream point cloud benchmarks.
\end{abstract}

%%%%%%%%% BODY TEXT
\section{Introduction}
\label{sec:intro}

\begin{figure}[!ht]
	\centering
	\includegraphics[width=0.48\textwidth]{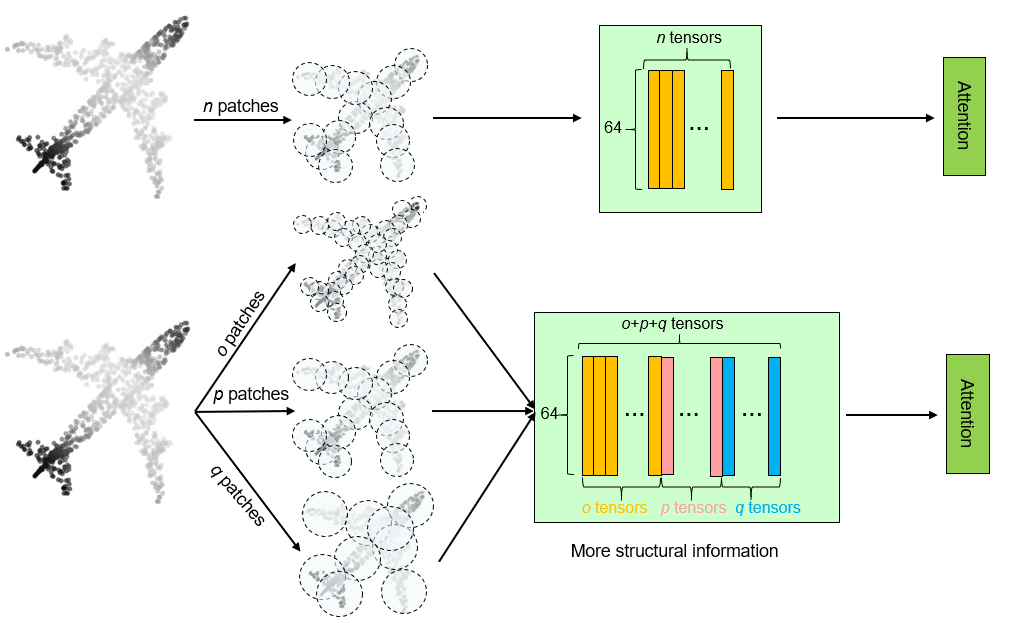}
	\caption{
		\label{fig_compareMGT}
Comparison between traditional point cloud Transformer and \emph{MGT}. \emph{MGT} contains more structural information. Top: Point Cloud Transformer. Bottom: \emph{MGT}. 
%, {\color{red} excluding normal vectors.
	}
\end{figure}

    Point cloud data is a widely used form of 3D data in fields such as autonomous driving  \cite{chen2017multi, shi2019pointrcnn, wang2023DuEqNet, fei2022Orthogonal}, augmented reality, and robotics \cite{rusu2008towards}. However, unlike traditional images, point cloud data has complex global and local structures that are non-Euclidean in nature, making extracting features in practical applications challenging.
    
    To address this challenge, researchers have proposed various deep learning-based 3D point cloud classification methods, which can be categorized into voxel-based methods \cite{maturana2015voxnet,shi2020pv}, multi-view methods \cite{kanezaki2018rotationnet,you2018pvnet}, and point set methods \cite{qi2017pointnet,qi2017pointnet++,wang2019dynamic, wang2022continual}. Voxel-based methods usually disrupt important spatial relationships in the measurement space. Multi-view methods project 3D cloud data into 2D images, resulting in the model's inability to fully capture geometric information and spatial relationships. PointNet \cite{qi2017pointnet} is a pioneer in point set methods, which uses spatial encoding for each point, such as multi-layer perceptrons (MLP) and pooling layers with shared weights, to collect point set features. Despite attempts to enhance these techniques in previous work \cite{thomas2019kpconv}, they still have limited ability to capture variations in local features. Therefore, all of these algorithms are unable to fully extract the complex geometric structure of point clouds for classification.
    
    Recently, the Transformer \cite{vaswani2017attention}, with a self-attention mechanism at its core, has been widely applied in natural language processing (NLP), image recognition, and object detection \cite{routeformer, detectdeformable, huang2019attention_AOA_iccv, dosovitskiy2020image, lan2023Couplformer}. Inspired by the tremendous success of Transformers in NLP and image domains, some researchers have attempted to introduce self-attention mechanisms into the point cloud classification field. For example, Point Cloud Transformer  \cite{zhao2021point, guo2021pct} uses self-attention mechanisms to capture the relationships between points and obtain global features. However, the aforementioned self-attention modules heavily rely on dot products in the Euclidean space.

\begin{figure}[!ht]
	\centering
	\includegraphics[width=0.5\textwidth]{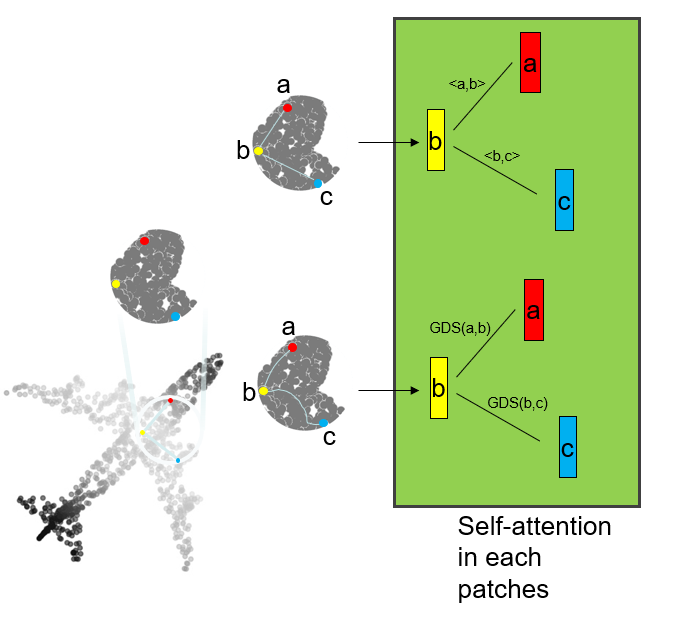}
	\caption{
		\label{fig_compareGDS}
Comparison between dot product-based self-attention and geodesic-distance-based self-attention on point cloud. Top: dot product-based self-attention. Bottom: geodesic-distance-based self-attention.}
\end{figure}

    In summary, the traditional data processing methods are no longer sufficient to handle the complex non-Euclidean structure of point cloud objects. Currently, a powerful feature extractor Transformer is needed to enhance the geometry of the point cloud and improve the local feature extraction capability while utilizing the Transformer's strong global feature acquisition ability. To this end, this paper proposes a novel Transformer, called the Multi-Scale Geometry-aware Transformer (\emph{MGT}), to extract complex geometric structures in point clouds for classification. \emph{MGT} divides the point cloud data into multiple patches with different numbers and sizes, thereby obtaining point cloud features from multiple scales. Figure \ref{fig_compareMGT} shows the comparison between traditional single-scale point cloud transformers and \emph{MGT}. Also, this self-attention method uses geodesic distance \cite{ligeodesic} instead of dot product multiplication in the self-attention module, which is more reasonable for processing point cloud data. Figure \ref{fig_compareGDS} provides a simple comparison between dot product attention and geodesic attention. Additionally, this paper proposes a Shared Local Feature Extractor (SLFE) module based on a sphere mapping algorithm to extract point cloud features using a feature extraction algorithm that is more suitable for point cloud data.

    The main contributions of this paper are summarized as:

    \begin{itemize}

    \item  \textbf{Multi-scale Patch Division Transformer}. \emph{MGT} divides point cloud data into multi-scale patches with different sizes, i.e., from small size to large size of patches, and feeds them into the transformer to explore the multiple scales of structures of point clouds.
    %capture the features of point clouds at multiple scales.

    \item  \textbf{Geometry-aware intra-patch representation}. This paper presents an SLFE module that enhances intra-patch local features and outputs a fixed-length vector for each patch. In the SLFE module, a novel operator, called sphere mapping, is presented to capture the local geometric structure of the neighbors of a patch, i.e., the angles between points in a patch.

    \item \textbf{Geometry-aware inter-patch representation}. We adopt a new self-attention mechanism based on computing geodesic distances to better capture the global features between patches.

    \end{itemize}

% \begin{itemize}

%     \item  \textbf{Multi-scale Patch Division Transformer}. \emph{MGT} divides point cloud data into patches of multiple scales and feeds them into the transformer to capture the features of point clouds at multiple scales.

%     \item  \textbf{Geometry-aware intra-patch representation}. This paper presents an SLFE module that enhances intra-patch local features and outputs a fixed-length vector for each patch.

%     \item \textbf{Geometry-aware inter-patch representation}. We adopt a new self-attention mechanism based on computing geodesic distances to better capture the global features between inter-patches.

% \end{itemize}

\begin{figure}[!ht]
	\centering
	\includegraphics[width=0.48\textwidth]{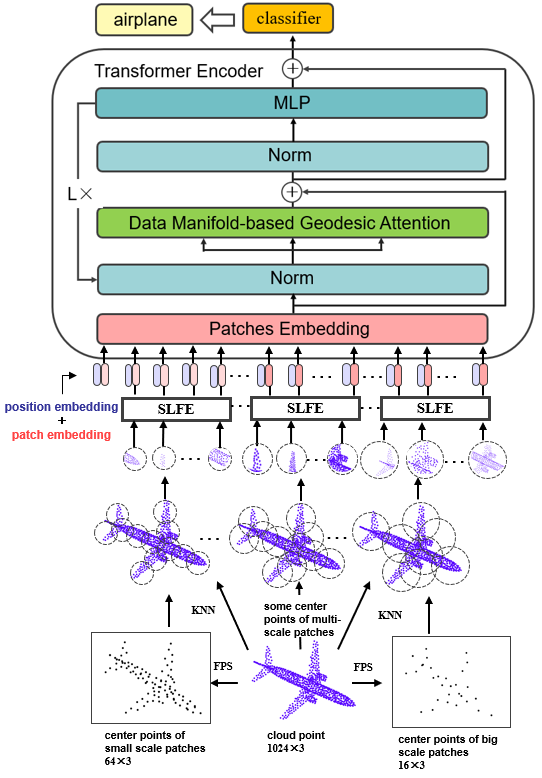}
	\caption{
		\label{fig_MGSA}
Model Overview. The Transformer Encoder architecture is adopted from ViT
and the input data shown in Figure 1 are three-dimensional coordinates.
%, {\color{red} excluding normal vectors.
	}
 
\end{figure}

%-------------------------------------------------------------------------
\section{Related Works}
\label{sec:formatting}

\subsection{Transformer in Visual field}

Recently, relying on powerful sequential data processing and
global features aggregation capabilities, Transformers have achieved great success in NLP and CV. 
% \cite{vaswani2017attention} introduced a Transformer architecture based on self-attention, without any recursive or convolutional components. 
% \cite{devlin2018bert} proposed BERT, a powerful two-way Transformer architecture in the NLP field that handles multi-modal tasks \cite{xu2021e2e}. 
Transformer \cite{vaswani2017attention}, with architecture based on self-attention, consists of no recursive nor convolutional components. 
BERT \cite{devlin2018bert} is a robust bidirectional Transformer architecture in the NLP field that handles multi-modal tasks \cite{xu2021e2e}. 
Researchers aimed to integrate point clouds and multi-view data using the Point View Network (PVNet) \cite{zhou2018voxelnet}. 
In \cite{ding2022cogview2}, a standard Transformer was applied to image patches based on pre-trained CNN models. These patches were then split into medium-sized tokens (words) for the Transformer's input, but the method had a high bias when there was no large training set. 

Self-attention networks are also used in 2D image processing due to the effectiveness of the Transformer model in NLP. \cite{wang2017residual} proposed a stack of attention and residual attention for image classification. \cite{ramachandran2019stand} applied self-attention to local image patches, aiming to develop a local self-attention layer that works on small and large inputs in vision models. \cite{zhao2020exploring} developed a family of self-attention operators, and ViT divides the image into patches of equal size, using them as an input sequence for the Transformer Encoder. Experiments show that ViT performs strongly with sufficient training data. 

\subsection{Transformer for Point Cloud Classification}

3D point clouds, a prevalent topic in computer vision, can be considered a special type of sequential data, making the use of Transformers a natural fit for point cloud task processing. Xie et al. \cite{xie2018attentional} proposed ShapeContextNet, which hierarchically constructs patches using a context method of convolution and uses a self-attention mechanism to combine the selection and feature aggregation processes into a training operation. Point2sequence \cite{liu2019point2sequence} starts with local areas and creates local features using LSTM-based attention modules, then uses a pooling method (referencing PointNet, Deep Sets \cite{zaheer2017deep}) to generate a global feature vector. Because it relies on the structure of sequences, this leads to higher computational complexity than the multi-head attention mechanism \cite{vaswani2017attention}.

Recent studies have discovered better methods to deal with point cloud data by Transformer.
Point Cloud Transformer \cite{zhao2021point, guo2021pct} of the first works to use Transformers for point cloud tasks, was recently made with a self-attention mechanism based on Edgeconv \cite{wang2019dynamic}. Point Cloud Transformer first increases the dimension of each point, then uses a Transformer to capture the relationships between points to obtain global features. Point-BERT \cite{yu2022point} designs a discrete Variational AutoEncoder (dVAE) for generating point tokens, inspired by \cite{devlin2018bert}, successfully introducing BERT-style \cite{lee2018pre} pre-training to point cloud tasks. Inspired by Masked Auto Encoders (MAE) \cite{he2022masked}, Point-MAE \cite{pang2022masked} further explores pre-training methods for point cloud Transformers, implementing a pre-training pipeline for point cloud tasks with a completely standard Transformer structure. These works lay a solid foundation for future Transformer use in point cloud tasks. 

In addition, some methods also use deep neural network structures and take the raw point cloud as their input. PointNet \cite{qi2017pointnet} is a pioneer of this type of method, which uses permutation-invariant operators (such as shared-weight MLPs and pooling layers) to aggregate features of a set of points. PointNet++ \cite{qi2017pointnet++} further explores this idea by introducing a hierarchical spatial structure that can capture the local features of point clouds. These methods are inspiring for extracting both global and local features from point clouds.

%%%%%%%%%%%%%%%%%%%%%%%%%%%%%%%%%%%%%%%%%%%%%%
\section{Method}
\label{sec:formatting}
%
%   In this section, a brief overview of the proposed MGSA framework is presented. 
   In this section, we introduce the \emph{MGT} model, an end-to-end Transformer framework that fully explores the geometric structures of point cloud objects. 
   At first, to examine the multiple scales of the structures of point clouds, 
   %in order to build a stronger Self-Attention for point clouds, 
  we split the data into multi-scale patches with different sizes, i.e., from small to large patches. 
	Then, we design a geometry-aware transformer model, 
%	discrimination learning 
	which explores two-level geometric structures, i.e., the Euclidean geometry of each intra-patch and the non-Euclidean geometry of inter-patches of point clouds. 

%
% In \emph{MGSA} model, we first describe the patches division of point clouds and then propose the strategy of extracting the geometry-aware features from intra-patches and inter patches individually. Finally
%
%Finally, we introduce the joint of two levels of features. 

%%%%%%%%%%%%%%%%%%%%%%%%%%%%%%%%%%%%%%%%%%%%%%
\subsection{The MGT Framework}
	With divided multi-scale patches, Figure~\ref{fig_MGSA} depicts the pipeline of the \emph{MGT} model, which assembles two basic modules. First, the \textit{intra-patch representation} module for geometry-aware feature extraction of each patch. Second, the \textit{inter-patch representation} module for learning manifold-based self-attention of multi-scale patches. The former extracts the local geometric characteristics and generates a fixed-length invariant representation vector for each patch, and the latter explores the non-Euclidean relation between multi-scale patches. The former is achieved by a developed local sharing feature extractor associated with a sphere mapping module, and a manifold-based self-attention module achieves the latter. 
   
       % This paper designs a lightweight local sharing feature extractor to share the same feature extractor of all the same size cloud patches of the same size, and enhance the extraction effect of the local features of the point cloud patch. 
    % %
    % This can effectively retain the local geometric characteristics of the point cloud patch when the encoding point cloud patch embedding to the point cloud patch.
% 	discriminative image representations.
%\subsection{Generate point cloud patches}

Before introducing the \emph{MGT} model, we first describe the patch division of point clouds.
\paragraph{Multi-scale Patch Division}
% preprocessing and    
\label{patchdiv}
In the first layer of the \emph{MGT} model, a multi-scale patch division is performed. 
As shown in Figure~\ref{fig_MGSA}, formally, given input point cloud data $X=\{x_1,x_2,...,x_N\} \in R^{N \times C_{in}}$ with $N$ points of each object, 
% with $N$ being the number of points of an object, % $X\in R^{N \times C_{in}}$, 
where $C_{in}=3$ when only using raw position coordinates, and $C_{in}=6$ when adding extra normal vector \cite{qi2017pointnet}. 

For applying the multi-scale patch division, it is critical to determine the center of each patch. 
%% \cite{wang2019dynamic}. 
Using the farthest point sampling (FPS), $S^\eta$ points are selected to treat as the center points, the set of which is denoted by $CT^\eta$, where $\eta$ denotes the multiple scales of patches. 
%K^\eta
%Each center point forms a patch 
Given a center point, the $K-$nearest neighbors algorithm (KNN) is used to select the nearest $K^\eta$ points 
% for each center point to get 
and form a point cloud patch $P^\eta$ with the size of $K^\eta\times C_{in}$. 
% The size of the patch $P^\eta$ 
%
Hence, $CT$ denotes the set of all patches
%contains the $\gamma$ kinds of the scales of patches, 
with multiple scales/sizes, from small to large. 
%
% with different sizes,  $CT^\eta$ contains the $\gamma$ kinds of the scales 
% some points are selected using farthest point sampling (FPS); only the selected points are preserved while others are directly discarded after this layer.
% First sampling $S^\eta$ center points $CT^\eta$ of small(sm) point cloud patches and large(lg) point cloud patches by Farthest Point Sampling (FPS), $\eta_\in{sm.lg}$. 

% According to these $ S^\eta$ center points, the KNN algorithm (KNN) is used to select the nearest $K^\eta$ points for each center point to get the point cloud patch $P^\eta$. 

The patch size $K^\eta$ varies according to the different implementation settings based on used datasets. 
For example, in a numerical experiment, we set $4$ kinds of scales/sizes for each patch, e.g., the patch size
$K^{\eta} = 32, 64, 128, 256$ with the kind of scales $\eta = 1,2,3,4$, and the corresponding number of patches $S^{\eta} = 64, 32, 16, 8$ with $S = \sum_{\eta = 1}^{4}S^{\eta}$.

% e.g., the patch size
% $K^{\eta_{\gamma}} = 32, 64, 128, 256$ with the kind of scales $\gamma = 1,2,3,4$, and the corresponding number of patches $S^{\eta_{\gamma}} = 64, 32, 16, 8$ with $S = \sum_{\gamma = 1}^{4}S^{\eta_{\gamma}}$.
% %
% % set $S^{sm}=64$, $K^{sm}=32$, $S^{lg}=16$, $S^{lg}=128$  . 

Finally, for the $\eta^{th}$ scale of point cloud patches, the patch centers and the patches can be denoted by:
\begin{equation}
\begin{split}
    CT^\eta &= \text{FPS}(X),\  CT^\eta\in R^{S^\eta \times C_{in}}, \\
      P^\eta &= KNN(K^\eta,X, CT^\eta),\  P^\eta\in R^{S^\eta \times K^\eta \times C_{in}}.
\end{split}
\end{equation}
%
% \begin{equation}
%   CT^\eta =\text{FPS}(X),  CT^\eta\in R^{S^\eta \times C_{in}}
%   \label{eq:important}
% \end{equation}

% \begin{equation}
%   P^\eta =KNN(K^\eta,X, CT^\eta),  P^\eta\in R^{S^\eta \times K^\eta \times C_{in}}
%   \label{eq:important}
% \end{equation}

% There are 
% Between point cloud patches, some areas may overlap. In addition, the coordinates of the point of the local area are converted into a local coordinate system relative to the heart point, which will have better convergence.

\begin{figure}[!ht]
	\centering
	\includegraphics[width=0.475\textwidth]{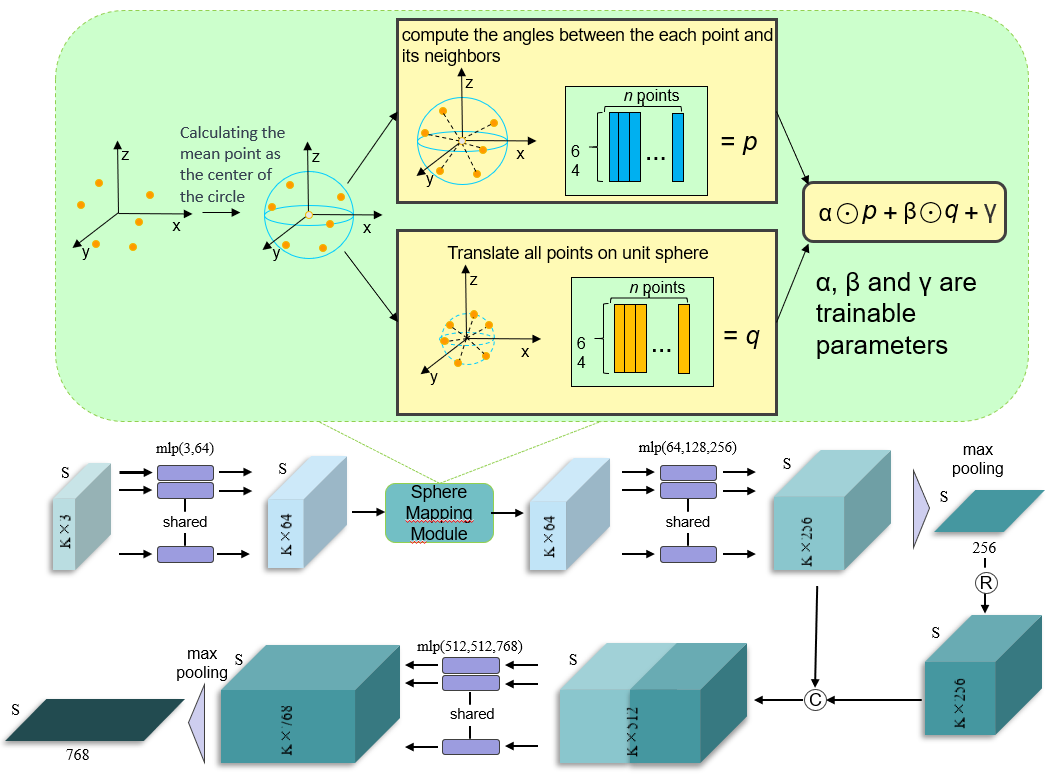}
	\caption{
		\label{fig_SLFE}
		Overview of the proposed \emph{SLFE} pipeline and the sphere mapping module. 
% 		\emph{SLFE} architecture diagram. 
The input data 
% shown in the figure 
are three-dimensional coordinates, without normal vectors. $S^{\eta}, K^{\eta}$ are abbreviated as $S$ and $K$, respectively. 
`®' represents repeat, `©' represents splicing. 
	}
\end{figure}%超页数了，改为单排版，页数够了再调回来
%

%\subsection{Locally shared feature extractor, SLFE}
\subsection{Geometry-aware Intra-Patch Representation} %using A Shared Local Feature Extractor with Sphere Mapping}
While dealing with patches with varying scales, a critical task is to design a feature extractor that generates a fixed-length representation for each patch. To address such an issue, we develop a unified Shared Local Feature Extractor (SLFE) to extract the fixed-length geometry-aware features for all scales of patches. 
%
% To obtain the feature (semantic information) of each point cloud patch, we try to use PointNet directly as the layout feature extractor. 
% But a shared PointNet model is difficult to achieve this goal, because the geometric structures of different local areas may require different feature extractors. 
% This paper draws on the idea of PointNet and designed a local shared feature extractor SLFE (Figure 2).
% Review of PointNet

\subsubsection{The \emph{SLFE} Model}
Before describing the \emph{SLFE} module, we briefly review the PointNet \cite{qi2017pointnet}.  
Formally, given a disordered point patch $X^\eta =\{x_1,x_2,...,x_{K^\eta}\} \in R^{K^\eta \times C_{in}}$,
PointNet regards $f: X^\eta \longmapsto R$ as a function embedding point set to vector,% as follows,
% define a function from point set to vector $f:X^\eta \longmapsto R$:
\begin{equation}
  f(\{x_1,x_2,...,x_n\})=\gamma(\max_{i=1,...,K^\eta} \{h(x_i)\})
  \label{eq:set_map_2_vector}
\end{equation}
where 
% Usually
$\gamma$ and $h$ are usually chosen as MLP networks to map the point cloud features into a high dimensional space for further processing, %increase the dimension of the points \cite{qi2017pointnet}, 
e.g., from $N\times 3$ to $N\times 1024$ in PointNet, where $N$ denotes the number of points. 
%
% The mapping function in Eq.~\eqref{eq:set_map_2_vector}
% can be interpreted as a point's spatial encoder \cite{qi2017pointnet++}, with unchanged order of input points. %
This mapping function in Eq.~\eqref{eq:set_map_2_vector}
could be interpreted as a point's linear spatial encoder \cite{qi2017pointnet++}, while ignoring the nonlinear geometric structure of the point clouds. 
%with unchanged order of input points.
%keeping
% is arranged unchanged on the input point, which 

Inspired by the %dimension increase of 
PointNet and its variants, we develop the \emph{SLFE} module, shown in Figure~\ref{fig_SLFE}, 
%
%As for the \emph{SLFE} model, Our contribution 
which can be concluded from two aspects: 
(1) We propose a novel local geometry extractor, sphere mapping, to extract the geometry-aware structures of patches. 
(2) A pipeline called \emph{SLFE} is proposed. Besides the sphere mapping, the $h$ used in this work contains the MLP as Eq.~\ref{eq:set_map_2_vector} of the PointNet and the MRC (MaxPooling, Repeat, Concat) module. 
Intuitively, as shown in Figure~\ref{fig_SLFE}, a feature aggregation of $S^\eta$ patches in the mid-term, i.e., the MaxPooling %为了与上文统一
of the $K^\eta$ neighbors of each point from a patch, makes the features of a patch more prominent and its semantic information easier to be captured.

% Compared with Eq.~\ref{eq:set_map_2_vector} of the PointNet, 
% includes not only MLP and geometric diffraction modules, 
%

% can 
% enhance the ability of local feature extraction, as well as semantic information
% %
% make the semantic information more prominent and easier to be captured. 
% %
% Stringing is conducive to enhancing the local features of each point of cloud patches and making local semantic information more vivid.

%%%%%%%%%%%%%%%%%%%%%%%%%%%%%%%%%%%%%%%%%%
%%%%%%%%%%%%%%%%%%%%%%%%%%%%%%%%%%%%%%%%%%
\paragraph{Sphere Mapping}
Since the point cloud has strong non-Euclidean geometric properties, its geometric features are challenging to capture correctly in Euclidean space. In the \emph{SLFE} module, we propose a novel sphere mapping, as shown in Figure~\ref{fig_SLFE}, to map the point cloud features to a sphere space for a better geometric analysis.
%In the \emph{SLFE} module, sphere mapping, as shown in Figure~\ref{fig_SLFE}, is proposed to extract the geometric properties of an input patch. 
%points set. 
%
%As shown in the formula (3), in the original PointNet, only one simple MLP network is used by  .  As shown in Figure 2, we % added a geometric diffraction module to the \emph{SLFE} architecture,
%
Formally, we use $\{P^\eta_{i,j}\}_{j=1,...,K^{\eta}}\in R^{K^{\eta\times d}}$ to represent the $K^\eta$ neighbor point of the ${{P_{i=1,...,S^\eta}^\eta}}\in R^d$, and each neighbor point ${P_{i,j}^\eta}$ is the vector with $d=64$. We then use the following formula to transform neighbor points to a sphere:
% %
% \begin{equation}
%   % \Vec{}
%   {\overline{P}^\eta_{i}}=\frac{1}{K^\eta} \sum_{j=1}^{K^\eta}  P^\eta_{i,j}
%   \label{eq:important}
% \end{equation}
% %
% \begin{equation}
%   \sigma^\eta = \sqrt{\frac{1}{K^{\eta}\times S^{\eta} \times d} \sum_{i=1}^{S^\eta} 
%   \sum_{j=1}^{K^\eta} (P^\eta_{i,j}  -  {\overline{P}^\eta_{i}} )^2 }
%   \label{eq:important}
% \end{equation}
%
\begin{align}
    {\overline{P}^\eta_{i}} & =\frac{1}{K^\eta} \sum_{j=1}^{K^\eta}  P^\eta_{i,j} \\
  \{ P^\eta_{i,j}\} & =  \alpha^\eta \odot \frac{  P^\eta_{i,j} - {\overline{P}^\eta_{i}} }{|P^\eta_{i,j}  -  {\overline{P}^\eta_{i}}| + \epsilon}  \\ 
   & + \frac{\beta^\eta}{K^\eta} \odot \sum_{s=1}^{K^\eta} \frac{ ( P^\eta_{i,j} - {\overline{P}^\eta_{i}}) ( P^\eta_{s,j} - {\overline{P}^\eta_{i}}) }{|P^\eta_{i,j}  -  {\overline{P}^\eta_{i}}| \cdot |P^\eta_{s,j}  -  {\overline{P}^\eta_{i}}| + \epsilon} + b^\eta
%   
%   \frac{ \{ P^\eta_{i,j}\} }{\|P^\eta_{i,j}  -  {\overline{P}^\eta_{i}}\| + \epsilon}+ \theta^\eta
  \label{eq:sphere_mapping}
\end{align}
where $\alpha^\eta,\beta^\eta, b^\eta \in R^d$  are learnable parameters, $\odot$  is Hadamard product, and $\epsilon =1e^{-5}$ is a stability coefficient to prevent the denominator to be $0$. 
In Eq.~\eqref{eq:sphere_mapping}, the first term is in charge of mapping the points to a uniform sphere, the second term computes the angles between the $i^{th}$ point and its neighbors, and the third term $b^\eta$ is the bias. 
%
% Note that $\sigma^\eta$ is a scalar that describes the feature deviation of all local areas and channels.
%

Through the sphere mapping module, the local geometric structure of the neighbors of a patch is mapped to a sphere. Hence we can extract the geometric relations (angles) between points in a more efficient way.
Therefore, the geometric characteristics of a patch are effectively extracted. 
\subsection{Geometry-aware Inter-Patch Representation} %using Manifold-based Attention}
With multi-scale patches and their features at hand, we employ a manifold-based geodesic Transformer to extract the non-Euclidean relations from inter-patches. 
%multi-scale 

\subsubsection{From Patch Encoding to Embedding}
\label{positionEmbedding}

Obtaining the multi-scale geometric patch representations $P^\eta \in R^{S^\eta \times K^\eta \times C_{in}}$, an embedding function $E_{P^\eta}=SLFE^\eta(P^\eta)$ is performed and then we concatenate the multi-scale embedded patches $E_p=Concat(E_{P^{sm}}, E_{P^{lg}}) \in R^{S \times d_{out}}$ for geodesic self-attention computation.

% the embedded features $E_{P^\eta}\in R^{S^\eta \times d_{out}}$ are available for geodesic self-attention computation. 

%Input all the point cloud patches $P^\eta \in R^{S^\eta \times K^\eta \times C_{in}}$ of different sizes in \ref{patchdiv} into and get the point cloud embedding $E_{P^\eta}\in R^{S^\eta \times d_{out}}$, that is, $E_{P^\eta}=SLFE^\eta(P^\eta)$. After that, we embedded the large cloud patch into the small point cloud patch for stitching to get the entire point cloud embedding $E_p=Concat(E_{P^{sm}}, E_{P^{sm}}) \in R^{S \times d_{out}}$, where $S=S^{sm}+S^{lg}=64+14=80$.

\subsubsection{Class Tag and Position Encoding}
Before the self-attention computation, we concatenate the class tag and implement the position encoding to the embedded features for better convergence. Similar to the claims in BERT\cite{devlin2018bert} and ViT\cite{dosovitskiy2020image}, we randomly initialize a learning-available class token $E_0 \in R^{d_{out}}$, and then stitch it with point cloud embedding $E_p$ to get the total sequence $E \in R^{(S+1)\times d_{out}}$ .

In the original Transformer, a position encoding module is applied to represent the order in natural language, which can reflect the position relationship between words. In this paper, in order to reflect the position relationship between the point cloud patches, we also add position encoding to the point cloud embedded to retain the location information. Considering that the point cloud data itself has location information, we use the center point coordinates of each point cloud patch to represent the location information of each point cloud patch. Specifically, we randomly initialize the class tag to $CT_{cls}\in R^3$ and then stitch with the central point coordinate of each point cloud patch, and then use a learned MLP layer to map the central point coordinate to the embedded dimension ($d_{out}$), then get the position encoding $E_{pos}\in R^{(S+1)\times d_{out}}$.

\subsubsection{Transformer Encoder}
%geodesic transformer of multi-scale patches
Following the pipeline of Transformer, the modified Transformer encoder is applied for feature encoding. The Transformer encoder includes \emph{MGT} with adjustable levels of patches. To extract the non-Euclidean inter-patch relationship, we implement the manifold-based geodesic self-attention to modify the self-attention module. We assume that for the point cloud, a typical non-Euclidean data, we should utilize the geodesic distance between its feature points rather than the inner product in Euclidean space to capture its relative relationship. For the input sequence $z$, we project the original features in Euclidean space onto an oblique manifold \cite{absi:book08}, i.e., a unit length Spheres denoted by $\mathcal{O M}$, and compute the geodesic self-attention of the patch features embedded on the oblique manifold. All concepts on geodesic and manifold are referred to \cite{absi:book08, hawe2013separable_cvpr}. The projection function $Proj( \cdot )$ can be described as:
\begin{equation}
    \begin{aligned}
        \boldsymbol{P}:=\mathrm{Proj}\left(\boldsymbol{z}\right)=\operatorname{Cat}\left(\left\{\frac{z_{i}}{\left\|z_{i}\right\|}\right\}_{i=0}^{L}\right),
    \end{aligned}
\end{equation}
where $Cat( \cdot )$ denotes concatenate function and $\| \cdot  \|$ denotes the square Frobenius norm in the ambient space. 

After the projection, the geodesic distance of an input pair of points $\{\boldsymbol{Q},\ \boldsymbol{K}\}$ on $\mathcal{O M}$ can be calculated as:
\begin{equation}
\label{eq_geodesic}
    \begin{aligned}
        \operatorname{dist}(\boldsymbol{Q}, \boldsymbol{K})=\sqrt{\sum_{i=1}^{n} \arccos ^{2}\left(\operatorname{diag}\left(\boldsymbol{Q}^{T} \boldsymbol{K}\right)\right)_{i}}\ ,
    \end{aligned}
\end{equation}

Therefore, the output of the \emph{MGT} module $z_{out}$ can be calculated as:
\begin{equation}
    z_{out} = softmax \left( - GDS(z)\right) \cdot W_v(z) ,
\end{equation}
where $GDS(\cdot)$ is the geodesic distance computation function defined in Eq.~\eqref{eq_geodesic} and $W_v$ is a linear layer.

Then, we input the sequence $z_0$ with the position encoding obtained in Section \ref{positionEmbedding} into the aforementioned Transformer Encoder with $L$ layers
% with the $L$ layer. 
(LN means LayerNorm). The specific process is as follows: 
\begin{equation}
  \begin{aligned}
  z_l&=MGT(LN(z_{l-1}))+z_{l-1},\ l=1,...,L \\
  z_l&=MLP(LN(z_l))+z_l,\ l=1,...,L
  \label{eq:important}
  \end{aligned}
\end{equation}
\begin{equation}    
  y=LN(z_{L}^0),\ y \in R^{d_{out}}
\end{equation}

After above steps, the class token $y$ with global classification information is obtained in the first element of the output sequence.
%Finally, the first element of the output sequence that captures the global class label $y$ is obtained, and the classification result is obtained by inputting it into the classification header.

%%%%%%%%%%%%%%%%%%%%%%%%%%%%%%%%%%%%%%%%%%%%%%%%%%
\section{Experiments}
\label{sec:Experiments}

In this section, we evaluate the multi-class classification performance of our proposed \emph{MGT} framework on several benchmarks.

In the ablation studies, we perform both qualitative and quantitative assessments of the \emph{MGT} framework's effectiveness. 
Some additional experiments are illustrated in supplementary materials.

\subsection{Experimental Environment and Configuration}

In training, we use the label smooth cross-entropy loss function \cite{muller2019does}. The batch size and the number of the epoch are set as $32$ and $250$, respectively. 

The SGD optimizer and CosineAnnealingLR are used to adjust the learning rate, and the initial learning rate is $0.02$. The configuration of the machine is shown in Table \ref{tab1:machine_configuration}.

Moreover, the overall accuracy (OA) and class-wise mean accuracy (mAcc) are used as the performance evaluation criteria for classification. 

Unless explicitly stated in the experimental study, we input $1024$ (1K) points and do not adopt the normal vector as the original point feature.

\begin{table}
  \centering
  \begin{tabular}{cc}
    \toprule
    Equipment & Name \\
    \midrule
    Operation System & Linux\\
    GPU & NVIDIA GEFORCE RTX 2080Ti(8) \\
    Framework & Pytorch1.8.1+cu101\\
    Language & Python3.7\\
    \bottomrule
  \end{tabular}
  \caption{ Experimental environment and configuration. Each group of experiments only uses one GPU, and eight are only comparative tests of multiple groups.}
  \label{tab1:machine_configuration}
\end{table}

\subsection{3D Object Classification}
The \textit{Modelnet40} \cite{wu20153d} dataset contains $40$ categories of $12311$ grid CAD models, including $9843$ training models and $2468$ test models.
For each data model, the sample of the grid surface $N = 1024$ points %and uses 
with normalized coordinates being the input of the network. 
During the training process, in order to improve the diversity of the sample, the following operations are adopted to enrich the data: 1) Randomly zoomed in the internal scale [$-0.8$, $1.25$]; %internal standards ;
2) Gaussian noise with the mean of $0$ and the variance of $0.02$ to shift each point Location; 3) Randomly discard some points with not exceeding $12.5\%$ of the total number of points.

\textit{ScanObjectNN} \cite{uy2019revisiting} dataset is a real-world dataset. The sample contains not only the target itself, but also background information and noise interference. 
Therefore, this dataset is more real, complex, and challenging. \textit{ScanObjectNN} contains $15,000$ objects, divided into $15$ and $2902$ unique object instances. In training, we use the same data augmentation methods as \textit{Modelnet40}.

% SHREC15 is a non-rigid shape data set with 1,200 shapes of 50 categories. The shape in SHREC15 is embedded in the 2D surface in 3D space. During the experiment, we randomly selected 17 three-dimensional models from each category as a training samples, and the rest were used as test samples. We are used as a random sampling of 1024 points from each shape. We use the same data enhancement method as we used in Modelnet40 and ScanObjectNN.

% To verify the effectiveness of the proposed method, we will compare it with some mainstream methods, such as Pointnet \cite{qi2017pointnet}, PointNet++\cite{qi2017pointnet++}, PointCNN \cite{li2018pointcnn}, PointWeb \cite{zhao2019pointweb}, SO-Net \cite{li2018so}, DGCNN \cite{wang2019dynamic}, PointConv \cite{wu2019pointconv}, SpiderCNN \cite{xu2018spidercnn}, PCT \cite{guo2021pct} and MLMSPT \cite{han2021point}. The results on \textit{ModelNet40} and \textit{ScanObjectNN} datasets are shown in Table \ref{tab:ModelNet40} and Table \ref{tab:ScanObjectNN}, respectively (xyz is the coordinate, n is the normal vector).

To verify the effectiveness of the proposed method, we will compare it with some mainstream methods. The results on \textit{ModelNet40} and \textit{ScanObjectNN} datasets are shown in Table \ref{tab:ModelNet40} and Table \ref{tab:ScanObjectNN}, respectively ($xyz$ is the coordinate, $n$ is the normal vector).

\begin{table}
  \centering
  \begin{tabular}{ccccc}
    \toprule
    Method &	Input &	\#Points	& OA\% &	mAcc\% \\
    \midrule
        Pointnet\cite{qi2017pointnet} & $xyz$ & 1K & $89.2$ & $86.2$ \\
        \multirow{2}*{PointNet++\cite{qi2017pointnet++}} & $xyz$ & 1K & $90.7$ & -  \\
        ~ & $xyz+n$ & 5K & $91.9$ & -  \\
        PointWeb\cite{zhao2019pointweb} & $xyz$ & 1K & $92.3$ & $89.4$  \\
        \multirow{2}*{SO-Net\cite{li2018so}} & $xyz+n$ & 2K & $90.9$ & $87.3$ \\
        ~ & $xyz+n$ & 5K & $93.1$ & $\textbf{90.9}$ \\
        PointConv\cite{wu2019pointconv} & $xyz$ & 1K & $92.5$ & - \\
        DGCNN\cite{wang2019dynamic} & $xyz$ & 1K & $92.9$ & $90.5$ \\
        PCT\cite{guo2021pct} & $xyz$ & 1K & $93.1$ & - \\
        MLMSPT\cite{han2021point} & $xyz$ & 1K & $\underline{92.9}$ & - \\
        \hline
        Ours & $xyz$ & 1K & $92.14$ & $90.38$ \\
        Ours & $xyz+n$ & 1K & $\textbf{93.19}$ & $\underline{90.45}$ \\
    \bottomrule
  \end{tabular}
  \caption{ Classification performance of mainstream methods on \textit{ModelNet40} dataset.The optimal results in the table are shown in $\textbf{bold}$, and the sub-optimal results are shown in $\underline{underline}$.}
  \label{tab:ModelNet40}
\end{table}

% \begin{table}
%   \centering
%   \begin{tabular}{ccc}
%     \toprule
%     Method &	Input & Accuracy\% \\
%     \midrule
%         Shape-DNA & xyz &  61.2  \\
%         SVM+HKS & xyz &  56.9   \\
%         PointNet & xyz &  69.4  \\
%         PointNet++ & xyz &  60.2   \\
%         Ours & xyz &  \textbf{69.42} \\
%     \bottomrule
%   \end{tabular}
%   \caption{This table represents the classification results obtained by several different methods in the case of using data set SHREC15. The optimal results in the table are shown in $\textbf{bold}$, and the sub optimal results are shown in $\textit{italic}$.}
%   \label{tab:example}
% \end{table}

\begin{table*}
  \centering
  \begin{tabular}{cccccccccc}
    \toprule
     % \diagbox{class}{algorithms} 
     &     3DmFV \cite{ben20183dmfv} &	Pointnet &	Pointnet++ &	SpiderCNN \cite{xu2018spidercnn} &	DGCNN  &	PointCNN \cite{li2018pointcnn} &	PointConv	 &	Ours \\
    \midrule
% bag & 39.8 & 36.1 & 49.1 & 43.4 & 59.4 & 57.8 & 52.6 & 67.9 & \textbf{68.1} \\
% bin & 62.8 & 69.8 & 84.4 & 75.9 & 82.4 & 82.9 & 84.2 & 83.4 & \textbf{85.7} \\
% box & 15.0 & 20.5 & 34.6 & 12.8 & 33.1 & 33.1 & 35.8 & 57.9 & \textbf{59.3} \\
% cabinet & 65.1 & 62.6 & 77.4 & 74.2 & \textbf{83.9} & 83.6 & 76.8 & 83.2 & 81.0 \\
% chair & 84.4 & 59.0 & 91.3 & 89.0 & 91.8 & 92.6 & 91.6 & 92.8 & \textbf{93.4} \\
% desk & 36.0 & 50.0 & 74.0 & 65.3 & 63.3 & 65.3 & 64.9 & \textbf{75.1} & 70.7 \\
% display & 62.3 & 73 & 79.4 & 74.5 & 77.0 & 78.4 & 79.7 & 85.1 & \textbf{89.3} \\
% door & 85.2 & 92.8 & 85.2 & 91.4 & 89 & 84.8 & 88.7 & \textbf{93.4} & 92.1 \\
% shelf & 60.6 & 72.6 & 72.6 & 78 & 79.3 & 84.2 & 85.1 & 87.1 & \textbf{88.2} \\
% table & 66.7 & 67.8 & 72.6 & 65.9 & \textbf{77.4} & 67.4 & 62.8 & 75.0 & 70.6 \\
% bed & 51.8 & 61.8 & 75.5 & 69.1 & 64.5 & 80.0 & 72.8 & 73.4 & \textbf{81.0} \\
% pillow & 61.9 & 67.6 & 81.0 & 80.0 & 77.1 & 80.0 & 79.5 & \textbf{89.2} & 88.4 \\
% sink & 46.7 & 64.2 & 80.8 & 65.8 & 75.0 & 72.5 & 73.9 & \textbf{81.3} & 80.0 \\
% sofa & 72.4 & 76.7 & 90.5 & 90.5 & 91.4 & 91.6 & 90.0 & 91.6 & \textbf{92.8} \\
% toilet & 61.2 & 55.3 & \textbf{85.9} & 70.6 & 69.4 & 71.8 & 73.5 & 85.4 & 82.9 \\
% \hline$
mAcc & $58.3$ & $63.9$ & $75.4$ & $69.8$ & $73.6$ & $75.1$ & $74.1$  & $\textbf{78.59}$ \\
OA &   $63.9$ & $68.2$ & $78.9$ & $73.7$ & $78.1$ & $78.5$ & $77.4$  & $\textbf{80.47}$ \\
    \bottomrule
  \end{tabular}
  \caption{ Classification performance of mainstream methods on ScanObjectNN dataset. The optimal results in the table are shown in \textbf{bold}}
  \label{tab:ScanObjectNN}
\end{table*}

The experimental results of Table \ref{tab:ModelNet40} and Table \ref{tab:ScanObjectNN} show that the proposed method has demonstrated very competitive results when compared with some mainstream methods. On the Modelnet40 dataset, the algorithms of this paper achieved $93.19\%$ and $90.45\%$ on the OA and mAcc indicators, respectively. 
%(using the method vector)
 The strongly competitive results on the two benchmark datasets show the effectiveness of the proposed method in comparison with the aforementioned baselines.

\subsection{Ablation Experiments}
In this paper, we use the SLFE to encode the point cloud into point cloud embedding. 
In the experiment, it is found that using the original PointNet++ as the local feature extractor is difficult to achieve ideal results. 
To this end, we draw on the ideas of PointNet, and develop a lightweight SLFE as a local feature extractor, which can have a better performance. 

In this section, we compare the experimental results by using the original PointNet and SLFE.

\begin{table}
  \centering
  \begin{tabular}{cccc}
    \toprule
    local feature extractor & \ Parameters &	OA\% &	mAcc\% \\
    \midrule
    PointNet &	$1.87$M &	$90.80$ &	$87.92$ \\
SLFE(Single) &	$0.31$M &	$93.31$ &	$90.63$ \\
    \bottomrule
  \end{tabular}
  \caption{ Comparison Experiment of feature extractor using PointNet and SLFE on MedelNet40}
  \label{tab:MedelNet40_abligation}
\end{table}

The experimental results in Table \ref{tab:MedelNet40_abligation} show that a single SLFE has only $0.31$M parameters, but the results by using SLFE are $2.51\%$ and $2.71\%$ higher than those using PointNet on OA and mAcc, respectively. 

This indicates that the developed SLFE  is effective as a local feature extractor. 
The ablation experiment, test results on ModelNet40 dataset, in Table \ref{tab:ModelNet40_module} discusses the necessity of SLFE in the modules.

\begin{table}
  \centering
  \begin{tabular}{ccccc}
    \toprule
    No. &	Sphere Mapping &	MRC &	OA\% &	mAcc\% \\
    \midrule
    $A$ &  $\color{red}\times$ & $\color{red}\times$ & $90.54$ & $87.64$ \\
$B$ & $\color{red}\times$ & $\color{green}\checkmark$ & $91.73$ & $88.34$ \\
$C$ & $\color{green}\checkmark$ & $\color{red}\times$ & $92.31$ & $90.37$ \\
$D$ & $\color{green}\checkmark$ & $\color{green}\checkmark$ & $93.31$ & $90.63$ \\
    \bottomrule
  \end{tabular}
  \caption{ Ablation experiments on \textit{ModelNet40}. $\color{red}\times$ indicates the experiment without the corresponding module, and $\color{green}\checkmark$ indicates the experiment with the corresponding module}
  \label{tab:ModelNet40_module}
\end{table}

According to the analysis of the ablation experiment results in the table, $A$ had the worst result without adding Sphere Mapping module and maximum pooling module.
Intuitively, in the middle term, feature aggregation is conducted for each point cloud. That is, the point features of each patch are pooled to the maximum, and the obtained local features are spliced with the features before aggregation to highlight the local features and make the local semantic information more distinct. 
The OA result in $B$ is $1.19\%$ higher than that in $A$, which indicates that this operation is effective and confirms our assumption. 
The OA result in $C$ is $1.77\%$ higher than that in $A$, and the Sphere Mapping module maps the local features to the unit length sphere, thus implicitly encoding the local geometric information and angles, making up for the lack of geometric information. So that all point clouds of the same size can better extract the local geometric features when sharing a local feature encoder. 
It is similar to the operation in $B$, that is, it highlights the local information and makes the local semantic information more distinct, which lays a good foundation for the later Transformer Encoder module to capture the relationship between local features. 
The results in $D$ show that $B$ and $C$ operations complement each other and make the algorithm obtain the optimal results.

\begin{table}
    \centering
        \begin{tabular}{cccccccc}
            \toprule 
                \multicolumn{2}{c}{}& \multicolumn{2}{c}{\textit{ModelNet40}}&\multicolumn{2}{c}{\textit{ScanObjectNN}}\\
                \multicolumn{2}{c}{}&OA\%&mAcc\%&OA\%&mAcc\%\\  
                    \hline 
                \multicolumn{2}{c}{Dot product}&$92.1$	&$89.38$&	$79.47$&	$76.57$\\  
                \multicolumn{2}{c}{Geodesic-distance}&$92.21$&	$89.96$&	$79.99$	&$76.82$\\
            \bottomrule
        \end{tabular}
    \caption{Experimental results on different self-attention methods}
    \label{tab:geodesic_or_not}
\end{table}

Also, in this section, we have discussed at length the impact of the geodesic-distance-based self-attention mechanism on the experimental results. In Table \ref{tab:geodesic_or_not}, we compared the experimental results of the traditional self-attention mechanism using dot product with the self-attention mechanism based on geodesic distance. The experimental results in the table show that the use of the geodesic distance-based self-attention mechanism improved the results on both the ModelNet40 and ScanobjectNN datasets. On the ModelNet40 dataset, using the geodesic-distance-based self-attention mechanism improved OA by $0.11\%$ and mAcc by $0.58\%$ compared to using the traditional dot product-based self-attention mechanism. For the ScanobjectNN dataset, OA was improved by $0.52\%$ and mAcc was improved by $0.25\%$. This result indicates that for point cloud data, using the geodesic distance on the manifold surface to express weights is more suitable than the traditional dot product.

\begin{table}
    \centering
        \begin{tabular}{cccccc}
            \toprule 
                \multicolumn{2}{c}{Number of scales}& \multicolumn{2}{c}{\textit{ScanObjectNN}}\\
                \multicolumn{2}{c}{}&OA\%&mAcc\%\\  
                    \hline 
                \multicolumn{2}{c}{1}&$79.68$	&$77.20$\\  
                \multicolumn{2}{c}{2}&$79.85$ &$77.52$\\
                \multicolumn{2}{c}{3}&$79.99$&$77.77$\\
            \bottomrule
        \end{tabular}
    \caption{Experimental results in different numbers of scales}
    \label{tab:num_of_scale}
\end{table}

In addition, we also explored the impact of the number of scales on classification accuracy. Table \ref{tab:num_of_scale} shows the different effects on experimental results by dividing point cloud data into different scales in Multi-scale Division. The ScanobjectNN dataset was used in the experiment. The results demonstrate that the multi-scale transformer performs better than the single-scale transformer. Dividing point cloud data into more scales allows more diverse data to enter the transformer, thus providing more point cloud structural information. The scale with more points but fewer patches provides more detailed global feature about the point cloud, while the scale with fewer points but more patches provides more detailed local feature .
\subsection{Point Number and Patch size}

% In this section, we will discuss the experimental results of point cloud data with different point numbers, as well as the impact of dividing the point cloud into patches of different sizes using multi-scale patch division on the experimental results. The experiment is based on the classification task on ScanobjectNN dataset.

In this section, we will discuss the experimental results of point cloud data with different point numbers. The experiment is based on the classification task on ScanobjectNN dataset.

\begin{table}
  \centering
  \begin{tabular}{ccc}
    \toprule
    Point Numbers & 	OA\% &	mAcc\% \\
    \midrule
    $512$ &		$78.99$ &	$76.08$ \\
    %$1024$ &		$79.57$ &	$77.56$ \\
    $1024$ &		$79.47$ &	$76.57$ \\
    $2048$ &		$79.71$ &	$77.27$ \\
    \bottomrule
  \end{tabular}
  \caption{ The effect of the number of points in a point cloud on the experimental results.}
  \label{tab:pointNumbers}
\end{table}

We conducted experiments on point cloud data with different numbers of points within the same dataset in Table \ref{tab:pointNumbers}. Each experiment used a different number of points extracted from the dataset for calculation. We conducted experiments on the ScanobjectNN dataset with $512$ points, $1024$ points, and $2048$ points, respectively. As shown in the table, the accuracy of \emph{MGT} gradually increased as the number of points in the dataset increased. 

% \begin{table}
%   \centering
%   \begin{tabular}{cccc}
%     \toprule
%     Patch numbers & Patch size & OA\% &	mAcc\% \\
%     \midrule
%     64 &  32 &	 \multirow{3}{*}{78.92} &	\multirow{3}{*}{75.9} \\
%     32 &  64 &	 & \\
%     16 & 128 &	 & \\
%         \midrule
%     64 &  32 &	 \multirow{3}{*}{77.99} &	\multirow{3}{*}{75.49} \\
%      8 & 256 &	& \\
%      4 & 512 &	& \\
%         \midrule    
%     32 &  64 &	 \multirow{3}{*}{76.75} &	\multirow{3}{*}{73.63} \\
%     16 & 128 &   & \\
%      8 & 256 &	& \\

%     \bottomrule
%   \end{tabular}
%   \caption{ Comparison Experiment in different scales.}
%   \label{tab:patchSize}
% \end{table}

% In section \ref{patchdiv}, we introduced the specific method of multi-scale patch division. The number of patches within each scale, and the number of points in each patch can be controlled for each point cloud data involved in the calculation. Table \ref{tab:patchSize} demonstrates the impact of patch size and patch number with different distributions on the experimental results. It can be observed that the experiment performs better when the patch size tends to be bimodal. This is because patches with different scales complement the global and local features.

\subsection{Transfer Learning}
The research of ViT point out that when the amount of training data is large enough, the global context modeling based on Transformer has stronger expression ability. Therefore, this paper conducts a migration learning experiment from a larger dataset to a smaller dataset.

ShapeNetCore v2 contains $55$ categories of $51300$ 3D models in total. 
We divide the entire data set into a training set and a verification set, but only perform pre-training on the training set. 
For each instance model, we use the FPS algorithm to uniformly sample $1024$ points as the point cloud input data. Note that, unlike the data enhancement in ModelNet40 and ScanObjectNN, we apply standard random scaling and random translation to enhance the data. 
During pre-training, Adam optimizer and CosineAnnexingLR are used to adjust the learning rate. 
The initial learning rate is set to $0.03$, weight\_decay is $0.005$, batchsize is $64$, and the epoch is $250$.
  
\begin{table}
\centering
\begin{tabular}{cccccccc}
\toprule 
\multicolumn{2}{c}{}& \multicolumn{2}{c}{\textit{ModelNet40}}&\multicolumn{2}{c}{\textit{ScanObjectNN}}\\
\multicolumn{2}{c}{}&OA\%&mAcc\%&OA\%&mAcc\%\\  
\hline 
\multicolumn{2}{c}{Direct training}&93.31	&$90.63$&	$82.81$&	$81.91$\\  
\multicolumn{2}{c}{Transfer Learning}&94.03&	$91.28$&	$83.52$	&$82.77$\\
\bottomrule
\end{tabular}
\caption{
Model performances with Transfer learning
\label{tal:transfer}
}
\end{table}

Compared with direct training, it can be seen from Table \ref{tal:transfer} that when a large dataset, ShapeNetCore v2, is used for pre-training first, and then fine-tuning on the ModelNet40 dataset, the accuracy of OA and mAcc of this method are $94.03\%$ and $91.28\%$, respectively. Fine-tuning in the ScanObjectNN dataset achieved $83.52\%$ and $82.77\%$ in accuracy, respectively. Therefore, the proposed method in this paper effectively inherits the excellent migration capability of Transformer architecture.

\subsection{Robustness Test and Analysis}
The experiments above verified that the proposed method is simple and effective in classification. 
This section will test the robustness of the model on data point loss. 
The index in this section is the OA of the model training conducted on the ModelNet40 training set (1K points), and then testing on the test set (the farthest point sampled with missing data).

\begin{figure}[!ht]
	\centering
	\includegraphics[width=0.5\textwidth]{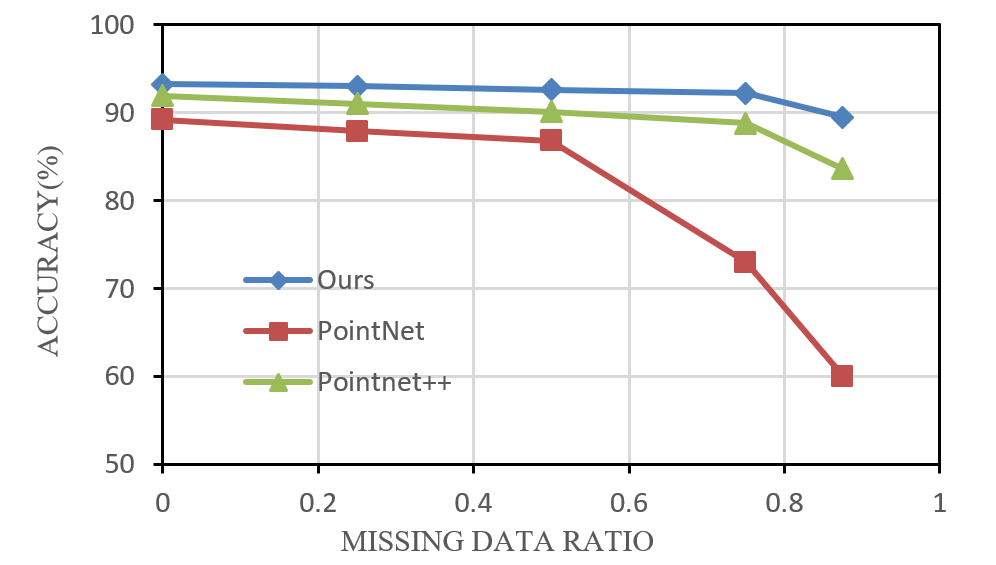}
	\caption{
		\label{fig_missing}
		Model performances with missing points.
	}
\end{figure}

In this paper, PointNet and Pointnet++, 
are selected as the comparative to test the robustness with point cloud loss (data with missing points). 
As shown in Figure \ref{fig_missing}, when the farthest point sampling is used to make the point cloud data of the test set lose $50\%$ (i.e. $512$ points are retained), the accuracy rate only decreases by $0.6\%$, which is lower than $1.8\%$ and $2.4\%$ of the decline of Pointnet++ and Pointnet. 
When the ratio of lost points is $87.5\%$ (that is, $128$ points are retained), the algorithm in this paper can still achieve an excellent accuracy of $89.5\%$, and Pointnet++ is $83.6\%$, while PointNet can only achieve an accuracy of $60\%$, with a significant decline in accuracy. 
This indicates that the proposed method has strong robustness on point cloud data with missing points. 
The experiment shows that,
when the recognition model can use the local information of the point cloud, it can be more robust to the loss of point cloud data (missing points). 
The proposed method can still have a high accuracy facing a high loss rate of point cloud data, owing to the full use of the rich local semantic information of point cloud and the excellent global modeling ability of Transformer Encoder architecture to capture the relationship between local semantic information.	

\section{Conclusion}
\label{sec:formatting}

In order to explore the complex geometric structures hidden in point clouds, this paper proposed a novel Transformer framework, \emph{MGT}, for the classification of point cloud objects. At first, we divide the data into multi-scale patches with different sizes, i.e., from small size to large size of patches, to explore the multiple scales of structures of point clouds. Then, we construct a geometry-aware transformer model, which leverages two-level geometric structures, i.e., the Euclidean geometry of each intra-patch and the non-Euclidean geometry of inter-patches of point clouds. The former is achieved by a local sharing feature extractor associated with a novel sphere mapping module, while the latter is accomplished using a manifold-based self-attention module. Compared with the mainstream methods, the accuracy of the proposed method on point cloud recognition has shown strong competitiveness and good robustness in the face of data point loss. In future,
%we consider extending the proposed method 
the proposed method can be extended to more point cloud processing tasks, 
such as object detection and scene segmentation.

\clearpage
%%%%%%%%% REFERENCES
{\small
\bibliographystyle{ieee_fullname}
\bibliography{egbib}
}

\end{document}

%%%%%%%%% TITLE
\title{Supplementary Material of Multi-scale Geometry-aware Transformer for 3D Point Cloud Classification}

\author{First Author\\
Institution1\\
Institution1 address\\
{\tt\small firstauthor@i1.org}
\and
Second Author\\
Institution2\\
First line of institution2 address\\
{\tt\small secondauthor@i2.org}
}

\maketitle
\ificcvfinal\thispagestyle{empty}\fi

%%%%%%%%% ABSTRACT
% \begin{abstract}
%    The ABSTRACT is to be in fully-justified italicized text, at the top
%    of the left-hand column, below the author and affiliation
%    information. Use the word ``Abstract'' as the title, in 12-point
%    Times, boldface type, centered relative to the column, initially
%    capitalized. The abstract is to be in 10-point, single-spaced type.
%    Leave two blank lines after the Abstract, then begin the main text.
%    Look at previous ICCV abstracts to get a feel for style and length.
% \end{abstract}

%%%%%%%%% BODY TEXT
\section{Classification on ModelNet10}

In this section, we provided a comparison with some mainstream 3D point cloud classification methods on the ModelNet10 \cite{wu20153d} dataset, as a supplement to Section 4.2 in the main text. 

The ModelNet10 dataset is a standard dataset for 3D shape recognition. It consists of 10 different object categories, each containing approximately 400 3D models, for a total of 4,899 3D models. The ModelNet10 dataset has been widely used to test and evaluate the performance of 3D shape recognition algorithms, such as point cloud classification, point cloud segmentation, 3D shape reconstruction, 3D object detection, and other tasks. It has become one of the standard datasets for evaluating 3D shape recognition algorithms. 

We compared several mainstream methods that use different 3D representation approaches for 3D shape classification on ModelNet10 in Table \ref{tab:ModelNet10}, including volumetric representation \cite{wu20153d,maturana2015voxnet,wang2019normalnet,wu2016learning,zhi2018toward,ma2017bv}, multi-view representation \cite{kasaei2020orthographicnet,han2018seqviews2seqlabels,shi2015deeppano}, and point-cloud representation \cite{riegler2017octnet,simonovsky2017dynamic,dominguez2018general,qi2017pointnet}, which is also used in proposed MGT. In addition, we also compared PolyNet \cite{yavartanoo2021polynet} with PolyShape representation method and VPC-CNN \cite{gezawa2022voxelized} with Voxelized Point Clouds representation method. 

The experimental setup and environment were based on Section 4.1 in the main text.

%-------------------------------------------------------------------------
% \subsection{Language}

% All manuscripts must be in English.

\begin{table}
  \centering
  \begin{tabular}{ccc}
    \toprule
        Method & Representation & Acc.(\%) \\
    \midrule
    3DshapeNet \cite{wu20153d} & Volumetric & $83.5$ \\
    VoxNet \cite{maturana2015voxnet} & Volumetric & $92$ \\
    NormalNet \cite{wang2019normalnet} & Volumetric & $93.1$ \\
    3D-GAN \cite{wu2016learning} & Volumetric & $91$ \\
    VSL \cite{liu2018learning} & Volumetric & $91$ \\
    lightNet \cite{zhi2018toward} & Volumetric & $93.3$ \\
    binVoxNetPlus \cite{ma2017bv} & Volumetric & $92.3$ \\
    OrthographicNet \cite{kasaei2020orthographicnet} & Multi-view & $88.5$ \\
    SeqView2seqlabels \cite{han2018seqviews2seqlabels} & Multi-view & $\underline{94.8}$ \\
    DeepPano \cite{shi2015deeppano} & Multi-view & $88.6$ \\
    PolyNet \cite{yavartanoo2021polynet} & PolyShape & $88.6$ \\
    VPC-CNN \cite{gezawa2022voxelized} & Voxelized Point Clouds & $93.4$ \\
    OctNet \cite{riegler2017octnet} & Point cloud & $90.4$ \\
    ECC \cite{simonovsky2017dynamic} & Point cloud & $90$ \\
    G3DNet \cite{dominguez2018general} & Point cloud & $93.1$ \\
    PointNet \cite{qi2017pointnet} & Point cloud & $77.6$ \\
 
    \midrule
    MGT(ours) & Point cloud & $\textbf{95.0}$ \\
    \bottomrule
  \end{tabular}
  \caption{ Classification performance of mainstream methods on \textit{ModelNet10} dataset. The optimal results in the table are shown in $\textbf{bold}$, and the sub-optimal results are shown in $\underline{underline}$.}
  \label{tab:ModelNet10}
\end{table}

{\small
\bibliographystyle{ieee_fullname}
\bibliography{egbib}
}